\documentclass[letterpaper]{article} % DO NOT CHANGE THIS
\usepackage{aaai23}  % DO NOT CHANGE THIS
\usepackage{times}  % DO NOT CHANGE THIS
\usepackage{helvet}  % DO NOT CHANGE THIS
\usepackage{courier}  % DO NOT CHANGE THIS
\usepackage[hyphens]{url}  % DO NOT CHANGE THIS
\usepackage{graphicx} % DO NOT CHANGE THIS
\usepackage{natbib}  % DO NOT CHANGE THIS AND DO NOT ADD ANY OPTIONS TO IT
\usepackage{caption} % DO NOT CHANGE THIS AND DO NOT ADD ANY OPTIONS TO IT
\usepackage{algorithm}
\usepackage{algorithmic}

\usepackage{multirow} 
\usepackage{amssymb}
\usepackage{pifont}
\usepackage{makecell}
\usepackage[capitalize]{cleveref}
\crefname{section}{Sec.}{Secs.}
\Crefname{section}{Section}{Sections}
\Crefname{table}{Table}{Tables}
\crefname{table}{Tab.}{Tabs.}
\usepackage{newfloat}
\usepackage{listings}
\DeclareCaptionStyle{ruled}{labelfont=normalfont,labelsep=colon,strut=off} % DO NOT CHANGE THIS
\floatstyle{ruled}
\newfloat{listing}{tb}{lst}{}
\floatname{listing}{Listing}
\title{CLIP-ReID: Exploiting Vision-Language Model for Image Re-Identification without Concrete Text Labels}
\author {
    Siyuan Li\textsuperscript{\rm 1},
    Li Sun\textsuperscript{\rm 1,\rm 2} \thanks{Corresponding author, email: sunli@ee.ecnu.edu.cn. },
    Qingli Li\textsuperscript{\rm 1}
}

\affiliations {
    \textsuperscript{\rm 1} Shanghai Key Laboratory of Multidimensional Information Processing\\
    \textsuperscript{\rm 2} Key Laboratory of Advanced Theory and Application in Statistics and Data Science\\
    East China Normal University, Shanghai, China
}

\begin{document}

\maketitle

\begin{abstract}
Pre-trained vision-language models like CLIP have recently shown superior performances on various downstream tasks, including image classification and segmentation. However, in fine-grained image re-identification (ReID), the labels are indexes, lacking concrete text descriptions. Therefore, it remains to be determined how such models could be applied to these tasks. This paper first finds out that simply fine-tuning the visual model initialized by the image encoder in CLIP, has already obtained competitive performances in various ReID tasks. Then we propose a two-stage strategy to facilitate a better visual representation. The key idea is to fully exploit the cross-modal description ability in CLIP through a set of learnable text tokens for each ID and give them to the text encoder to form ambiguous descriptions. In the first training stage, image and text encoders from CLIP keep fixed, and only the text tokens are optimized from scratch by the contrastive loss computed within a batch. In the second stage, the ID-specific text tokens and their encoder become static, providing constraints for fine-tuning the image encoder. With the help of the designed loss in the downstream task, the image encoder is able to represent data as vectors in the feature embedding accurately. The effectiveness of the proposed strategy is validated on several datasets for the person or vehicle ReID tasks. Code is available at https://github.com/Syliz517/CLIP-ReID.

\end{abstract}

\section{Introduction}
Image re-identification (ReID) aims to match the same object across different and non-overlapping camera views. Particularly, it focuses on detecting the same person or vehicle in the surveillance camera networks. ReID is a challenging task mainly due to the cluttered background, illumination variations, huge pose changes, or even occlusions. Most recent ReID models depend on building and training a convolution neural network (CNN) so that each image is mapped to a feature vector in the embedding space before the classifier. Images of the same object tend to be close, while different objects become far away in this space. The parameters of CNN can be effectively learned under the guidance of cross entropy loss together with the typical metric learning loss like center or triplet loss \cite{Triplet}. 

\begin{figure}[t]
\centering
\includegraphics[width=\linewidth,scale=1.00]{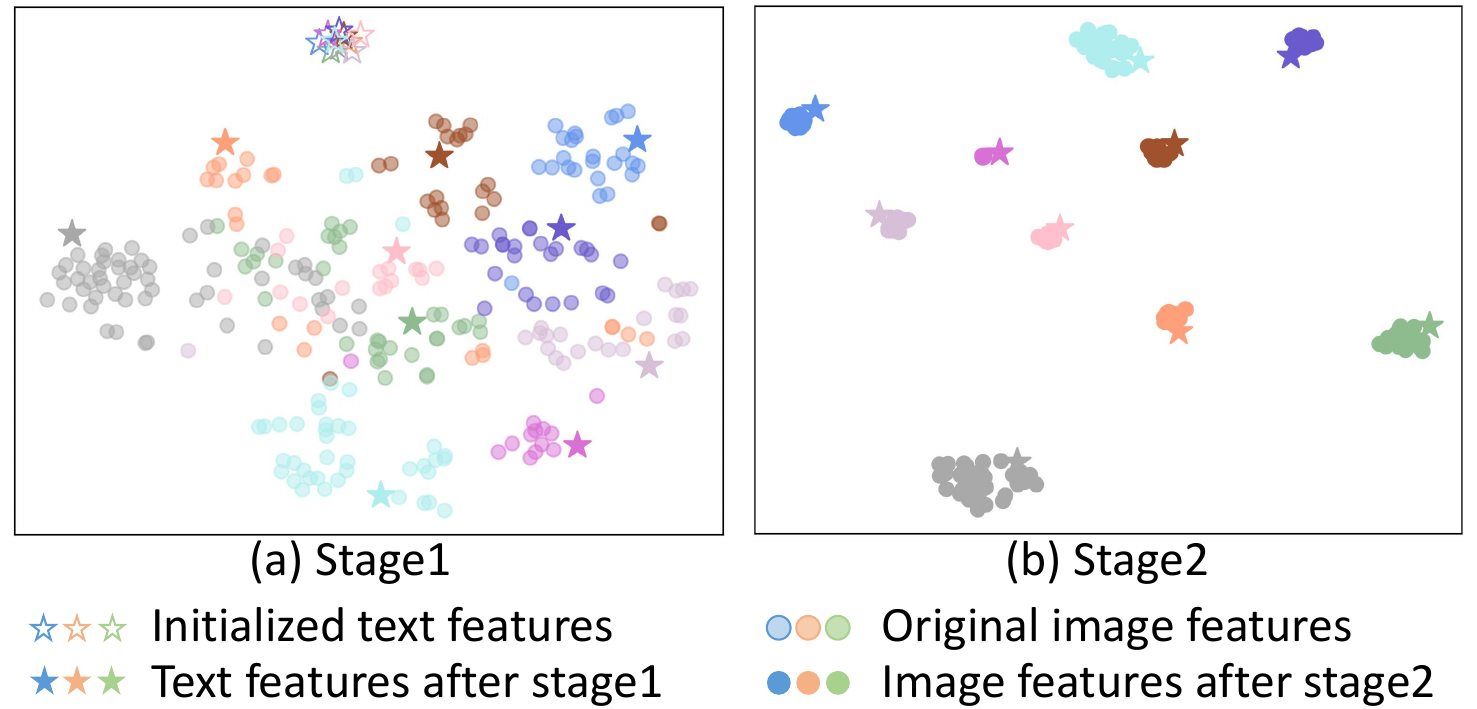}
\caption{t-SNE visualization on image and text features \cite{t-sne}. Randomly selected 10 persons in the MSMT17 are represented by different colors. The dots and pentagons indicate the image and text features, respectively. (a) and (b) show the data distributions after the first and second training stage.
}
\label{fg:t-sne}
\end{figure}

\begin{figure*}[t]
\centering
\includegraphics[width=\linewidth,scale=1.00]{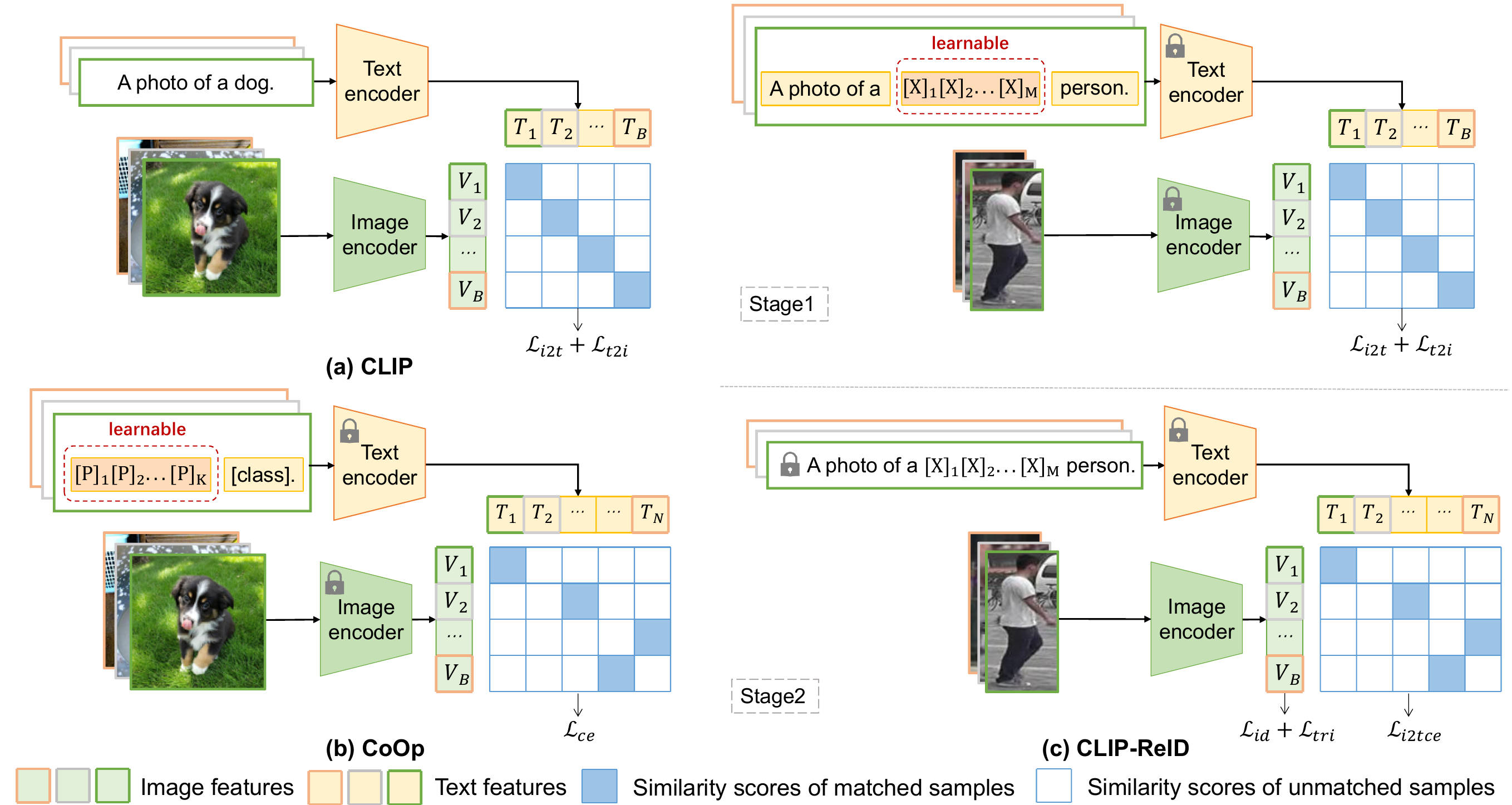}
\caption{Overview of our approach compared to CLIP and CoOp. (a) describes the model of CLIP, using pairs of text and image to train the image encoder and text encoder. (b) shows the model of CoOp, which fixes the image encoder and text encoder and fine-tunes text prompt in the downstream dataset. (c) is our proposed CLIP-ReID method, which fixes the text encoder and image encoder in the first training stage, optimizes a set of learnable text tokens to generate the text features, and then uses the text features to optimize the image encoder in the second training stage.}
\label{fg:method}
\end{figure*}

Although CNN-based models for ReID have achieved good performance on some well-known datasets, it is still far from being used in a real application. CNN is often blamed for only focusing on a small irrelevant region in the image, which indicates that its feature is not robust and discriminative enough. Recently, vision transformers like ViT \cite{ViT} have become popular in many tasks, and they have also shown better performances in ReID. Compared to CNN, transformers can model the long-range dependency in the whole image. However, due to a large number of model parameters, they require a big training set and often perform erratically during optimization. Since ReID datasets are relatively small, the potential of these models is not fully exploited yet. 

Both CNN-based and ViT-based methods heavily rely on pre-training. Almost all ReID methods need an initial model trained on ImageNet, %\cite{ImageNet}
which contains images manually given one-hot labels from a pre-defined set. Visual contents describing rich semantics outside the set are completely ignored. Recently, cross-modal learning like CLIP \cite{CLIP} connects the visual representation with its corresponding high-level language description. They not only train on a larger dataset %LAION-400 \cite{LAION-400M} 
but also change the pre-training task, matching visual features to their language descriptions. Therefore, the image encoder can sense a variety of high-level semantics from the text and learns transferable features, which can be adapted to many different tasks. \emph{E.g.}, given a particular image classification task, the candidate text labels are concrete and can be combined with a prompt, such as ``A photo of a", to form the text descriptions. The classification is then realized by comparing image features with text features generated by the text encoder, which takes the text description of categories as input. Note that it is a zero-shot solution without tuning any parameters for downstream tasks but still gives satisfactory results. Based on this, CoOp \cite{CoOP} incorporates a learnable prompt for different tasks. The optimized prompt further improves the performance.

CLIP and CoOp need text labels to form text descriptions in downstream tasks. However, in most ReID tasks, the labels are indexes, and there are no specific words to describe the images, so the vision-language model has not been widely adopted in ReID. In this paper, we intend to exploit CLIP fully. We first fine-tune the image encoder by directly using the common losses in ReID, which has already obtained high metrics compared to existing works. We use this model as our baseline and try to improve it by utilizing the text encoder in CLIP. A two-stage strategy is proposed, which aims to constrain the image encoder by generating language descriptions from the text encoder. A series of learnable text tokens are incorporated, and they are used to describe each ID ambiguously. In the first training stage, both the image and text encoder are fixed, and only these tokens are optimized. In the second stage, the description tokens and text encoder keep static, and they together provide ambiguous descriptions for each ID, which helps to build up the cross-modality image to text cross-entropy loss. Since CLIP has CNN-based and ViT-based models, the proposed method is validated on both ResNet-50 and ViT-B/16. The two types of the model achieve the state-of-the-art on different ReID datasets. Moreover, our method can also support the input of camera ID and overlapped token settings in its ViT-based version.

\cref{fg:t-sne} simultaneously visualizes image and text features in 2D coordinates, which could help to understand our training strategy. In the first stage, the text feature of each ID is adapted to its corresponding image features, making it become ambiguous descriptions. In the second stage, image features gather around their text descriptions so that image features from different IDs become distant. 

In summary, the contributions of this paper lie in the following aspects:

\begin{itemize}
    \item To our knowledge, we are the first to utilize CLIP for ReID. We provide competitive baseline models on several ReID datasets, which are the result of fine-tuning the visual model initialized by the CLIP image encoder.
    \item We propose the CLIP-ReID, which fully exploits the cross-modal describing ability of CLIP. In our model, the ID-specific learnable tokens are incorporated to give ambiguous text descriptions, and a two-stage training strategy is designed to take full advantage of the text encoder during training.
    \item We demonstrate that CLIP-ReID has achieved state-of-the-art performances on many ReID datasets, including both person and vehicle.
\end{itemize}

\section{Related Works}
\subsection{Image ReID} Previous ReID works focus on learning discriminative features like foreground histograms \cite{das2014consistent}, local maximal occurrences \cite{liao2015person}, bag-of-visual words \cite{market1501}, or hierarchical Gaussian descriptors \cite{matsukawa2016hierarchical}. On the other hand, it can also be solved as a metric learning problem, expecting a reasonable distance measurement for inter- and intra-class samples \cite{koestinger2012large}. %\cite{,pedagadi2013local}.
These two aspects are naturally combined by the deep neural network \cite{yi2014deep}, in which the parameters are optimized under an appropriate loss function with almost no intentional interference. Particularly, with the scale development of CNN on ImageNet, ResNet-50 \cite{ResNet} has been regarded as the common model \cite{strongbaseline} for most ReID datasets. 

Despite the powerful ability of CNN, it is blamed for its irrelevant highlighted regions, which is probably due to the overfitting of limited training data. OSNet \cite{OSNeT} gives a lightweight model to deal with it. Auto-ReID \cite{AutoREID} and CDNet \cite{CDNet} employ network architecture search for a compact model. OfM \cite{OFM} proposes a data selection method for learning a sampler to choose generalizable data during training. Although they obtain good results on some small datasets, performances drop significantly on large ones like MSMT17. 

Introducing prior knowledge into the network can also alleviate overfitting. An intuitive idea is to use features from different regions for identification. PCB \cite{PCB} and SAN \cite{SAN} divides the feature into horizontal stripes to enhance its ability to represent the local region. MGN \cite{MGN} utilizes a multiple granularity scheme on feature division to enhance its expressive capabilities further, and it has several branches to capture features from different parts. Therefore, model complexity becomes its major issue. BDB \cite{dai2019batch} has a simple structure with only two branches, one for global features and the other for local features, which employs a simple batch feature drop strategy to randomly erase a horizontal stripe for all samples within a batch. CBDB-Net \cite{CBDB-Net} enhances BDB with more types of feature dropping. Similar multi-branch approaches \cite{hardmix, LTReID, ALDER, PRN, PGAN, CFVMNet} %\cite{CAMA,hardmix,LTReID,ALDER,PRN,PGAN,CFVMNet}
with the purpose of mining rich features from different locations are also proposed, and they can be improved if the semantic parsing map participates during training \cite{SAN-person, ISP, PVEN, SPAN}. 

Attention enlarges the receptive field, hence is another way to prevent the model from focusing on small areas. In RGA \cite{RGA}, non-local attention is performed along spatial and channel directions. ABDNet \cite{ABD-Net} adopts a similar attention module and adds a regularization term to ensure feature orthogonality. HOReID \cite{HOReID} extends the traditional attention into high-order computation, giving more discriminative features. CAL \cite{CAL} provides an attention scheme for counterfactual learning, which filters out irrelevant areas and increases prediction accuracy. Recently, due to the power of the transformer, it has become popular in ReID. PAT \cite{PAT} and DRL-Net \cite{DRL-Net} build on ResNet-50, but they utilize a transformer decoder to exploit image features from CNN. In the decoder attention block, learnable queries first interact with key tokens from the image and then are updated by weighted image values. They are expected to reflect local features for ReID. TransReID \cite{transreid}, AAformer \cite{AAformer} and DCAL \cite{DCAL} all use encoder attention blocks in ViT, and they obtain better performance, especially on the large dataset. 

This paper implements both CNN and ViT models initialized from CLIP. Benefiting from the two-stage training, both achieve SOTA on different datasets. 

\subsection{Vision-language learning.} Compared to supervised pre-training on ImageNet, vision-language pre-training(VLP) has significantly improved the performance of many downstream tasks by training to match image and language. CLIP \cite{CLIP} and ALIGN \cite{ALIGN} are good practices, which utilize a pair of image and text encoders, and two directional InfoNCE losses computed between their outputs for training. Built on CLIP, several works \cite{li2022blip, ViLT} have been proposed to incorporate more types of learning tasks like image-to-text matching and mask image/text modeling. %ViLT \cite{ViLT} simplifies the processing of visual inputs to the same convolution-free manner as textual inputs. 
ALBEF \cite{ALBEF} aligns the image and text representation before fusing them through cross-model attention. SimVLM \cite{SIMVLM} uses a single prefix language modeling objective for end-to-end training. 

Inspired by the recent advances in NLP, prompt or adapter-based tuning becomes prevalent in vision domain
% Inspired by natural language processing (NLP), some works start to investigate prompt tuning and adapter to the field of vision. 
CoOp \cite{CoOP} proposes to fit in a learnable prompt for image classification. CoCoOp \cite{CoCoOp} learns a light-weight visual network to give meta tokens for each image, combined with a set of learnable context vectors. CLIP-Adapter \cite{CLIP-Adapter} adds a light-weight module on top of both image and text encoder.

In addition, researchers investigate different downstream tasks to apply CLIP. DenseCLIP \cite{rao2022denseclip} and MaskCLIP \cite{zhou2021denseclip} apply it for per-pixel prediction in segmentation. ViLD \cite{gu2021open} adapts image and text encoders in CLIP for object detection. EI-CLIP \cite{EL-CLIP} and CLIP4CirDemo \cite{CLIP4CirDemo} use CLIP to solve retrieval problems. However, as far as we know, no works deal with ReID based on CLIP.

\begin{algorithm}[t]
\caption{CLIP-ReID's training process.}
\label{alg:algorithm}
\textbf{Input}: batch of images $x_i$ and their corresponding texts $t_{y_i}$.\\
\textbf{Parameter}: a set of %ambiguous 
learnable text tokens $\rm[X]_m$($\rm m\in{1,...M}$) %$\rm[X]_1[X]_2[X]_3...[X]_M$ in $t_{y_i}$ 
for all IDs existing in training set $\mathcal X$, an image encoder $\mathcal I$ and a text encoder $\mathcal T$, linear layers $g_V$ and $g_T$.

\begin{algorithmic}[1] %[1] enables line numbers
\STATE Initialize $\mathcal{I}$, $\mathcal{T}$, $g_V$ and $g_T$ from the pre-trained CLIP. Initialize $\rm[X]_m$($\rm m\in{1,...M}$) randomly.
\WHILE {in the 1st stage}
    \STATE $s(V_i,T_{y_i}) = g_V(\mathcal I(x_i))\cdot g_T(\mathcal T(t_{y_i}))$
    \STATE Optimize  $\rm[X]_m$ by \cref{eq:stage1loss}.
\ENDWHILE
\FOR {$y_i=1$ to $N$}
    \STATE  $text_{y_i} = g_T(\mathcal T(t_{y_i}))$
\ENDFOR
\WHILE{in the 2nd stage}
    \STATE $s(V_i,T_{y_i}) = g_T(\mathcal I(x_i))\cdot text_{y_i}$
    \STATE Optimize $\mathcal I$ by \cref{eq:stage2loss}.
\ENDWHILE
\end{algorithmic}
\end{algorithm}

\section{Method}
\subsection{Preliminaries: Overview of CLIP}
We first briefly review CLIP. It consists of two encoders, an image encoder $\mathcal{I}(\cdot)$ and a text encoder $\mathcal{T}(\cdot)$. The architecture of $\mathcal{I}(\cdot)$ has several alternatives. Basically, a transformer like ViT-B/16 %\cite{ViT}
and a CNN like ResNet-50 %\cite{ResNet}
are two models we work on. Either of them is able to summarize the image into a feature vector in the cross-modal embedding.

On the other hand, the text encoder $\mathcal{T}(\cdot)$ is implemented as a transformer, %\cite{Attnisallyouneed}
which is used to generate a representation from a sentence. Specifically, given a description such as ``A photo of a $\rm[class]$." where $\rm[class]$ is generally replaced by concrete text labels. $\mathcal{T}(\cdot)$ first converts each word into a unique numeric ID by lower-cased byte pair encoding (BPE) with 49,152 vocab size \cite{BPE}. Then, each ID is mapped to a 512-d word embedding. To achieve parallel computation, each text sequence has a fixed length of 77, including the start $\rm[SOS]$ and end $\rm[EOS]$ tokens. After a 12-layer model with 8 attention heads, the $\rm[EOS]$ token is considered as a feature representation of the text, which is layer normalized and then linearly projected into the cross-modal embedding space.

Specifically, $i\in \{1...B\}$ denotes the index of the images within a batch. Let $img_i$ be the $\rm[CLS]$ token embedding of image feature, while $text_i$ is the corresponding $\rm[EOS]$ token embedding of text feature, then compute the similarity between $img_i$ and $text_i$:

\begin{equation}\label{eq:svt}
s(V_i,T_i) = V_i \cdot T_i = g_V(img_i)\cdot g_T(text_i)
\end{equation}
where $g_V(\cdot)$ and  $g_V(\cdot)$ are linear layers projecting embedding into a cross-modal embedding space. The image-to-text contrastive loss $\mathcal{L}_{i2t}$ is calculated as:
\begin{equation}\label{eq:li2t}
\mathcal{L}_{i2t}(i) = - \log\frac{\exp(s(V_i,T_i))}{\sum_{a=1}^B \exp(s(V_i,T_a))}
\end{equation}
and the text-to-image contrastive loss $\mathcal{L}_{t2i}$:
\begin{equation}\label{eq:lt2i}
\mathcal{L}_{t2i}(i) = - \log\frac{\exp(s(V_i,T_i))}{\sum_{a=1}^B \exp(s(V_a,T_i))}
\end{equation}
where numerators in \cref{eq:li2t} and \cref{eq:lt2i} are the similarities of two embeddings from matched pair, and the denominators are all similarities with respect to anchor $V_i$ or $T_i$.

For regular classification tasks, CLIP converts the concrete labels of the dataset into text descriptions, then produces embedding feature $T_i$ , $V_i$ and aligns them. CoOp incorporates a learnable prompt for different tasks while entire pre-trained parameters are kept fixed, as depicted in \cref{fg:method}(b). However, it is difficult to exploit CLIP in ReID tasks where the labels are indexes instead of specific text. 

\subsection{CLIP-ReID}
To deal with the above problem, we propose CLIP-ReID, which complements the lacking textual information by pre-training a set of learnable text tokens. As is shown in \cref{fg:method}(c), our scheme is built by pre-trained CLIP with the two stages of training, and its metrics exceed our baseline.

\subsubsection{The first training stage.} We first introduce ID-specific learnable tokens to learn ambiguous text descriptions, which are independent for each ID. Specifically, the text descriptions fed into $\mathcal{T}(\cdot)$ are designed as ``A photo of a $\rm[X]_1[X]_2[X]_3...[X]_M$ person/vehicle", where each $\rm[X]_m$($\rm m\in{1,...M}$) is a learnable text token with the same dimension as word embedding. $\rm M$ indicates the number of learnable text tokens. In this stage, we fix the parameters of $\mathcal{I}(\cdot)$ and $\mathcal{T}(\cdot)$, and only tokens  $\rm[X]_m$ are optimized. 

Similar to CLIP, we use $\mathcal{L}_{i2t}$ and $\mathcal{L}_{t2i}$, but replace $text_i$ with $text_{y_i}$ in \cref{eq:svt}, since each ID shares the same text description. Moreover, for $\mathcal{L}_ {t2i}$, different images in a batch probably belong to the same person, so $T_{y_i}$ may have more than one positive, we change it to: %there is more than one positive for $T_{y_i}$, we change it to:

\begin{equation}\label{eq:lt2i_P}
\mathcal{L}_{t2i}(y_i) = \frac{-1}{|P(y_i)|}\sum_{p\in P(y_i)} \log\frac{\exp(s(V_p,T_{y_i}))}{\sum_{a=1}^B \exp(s(V_a,T_{y_i}))}
% \mathcal{L}_{t2i}(y_i) = \frac{-1}{|P(y_i)|}\sum_{p\in P(y_i)} \log\frac{\sum_{p\in P(y_i)}\exp(s(V_p,T_{y_i}))}{\sum_{a=1}^B \exp(s(V_a,T_{y_i}))}
\end{equation}
Here, $P(y_i)=\{p\in {1...B}:y_p = y_i\}$ is the set of indices of all positives for $T_{y_i}$ in the batch, and $|\cdot| $ is its cardinality.

By minimizing the loss of $\mathcal{L}_{i2t}$ and $\mathcal{L}_{t2i}$, the gradients are back-propagated through the fixed $\mathcal{T}(\cdot)$ to optimize $\rm[X]_1[X]_2[X]_3...[X]_M$, taking full advantage of $\mathcal{T}(\cdot)$.
\begin{equation}\label{eq:stage1loss}
\mathcal L_{stage1} = \mathcal{L}_{i2t} + \mathcal{L}_{t2i} 
\end{equation}

To improve the computation efficiency, we obtain all the image features by feeding the whole training set into $\mathcal{I}(\cdot)$ at the beginning of the first training stage. For a dataset with $N$ IDs, we save $N$ different $T_{y_i}$ of all IDs at the end of this stage, preparing for the next stage of training.

\subsubsection{The second training stage.}In this stage, only parameters in $\mathcal{I}(\cdot)$ are optimized. To boost the final performance, we follow the general strong pipeline of object ReID \cite{strongbaseline}. We employ the %commonly used 
triplet loss $\mathcal{L}_{tri}$ and ID loss $\mathcal{L}_{id}$ with label smoothing for optimization, they are calculated as: %. The $\mathcal{L}_{id}$ and $\mathcal{L}_{tri}$ are calculated as:

\begin{equation}\label{eq:lid}
\mathcal{L}_{id} = \sum_{k=1}^N -q_k\log(p_k)
\end{equation}
\begin{equation}\label{eq:ltri}
\mathcal{L}_{tri} = \max(d_p-d_n+\alpha,0)
\end{equation}
where $q_k = (1-\epsilon)\delta_{k,y}+\epsilon/N$ denotes value in the target distribution, and $p_k$ represents ID prediction logits of class $k$, $d_p$ and $d_n$ are feature distances of positive pair and negative pair, while $\alpha$ is the margin of $\mathcal{L}_{tri}$.

To fully exploit CLIP, for each image, we can use the text features obtained in the first training stage to calculate image to text cross-entropy $\mathcal{L}_{i2tce}$ as is shown in \cref{eq:li2tce}. Note that following $\mathcal{L}_{id}$, we utilize label smoothing on $q_k$ in $\mathcal{L}_{i2tce}$. %Since all IDs belong to the same category so there is duplicated information between different text features, we propose label smoothing in $\mathcal{L}_{i2tce}$:
\begin{equation}\label{eq:li2tce}
\mathcal{L}_{i2tce}(i) = \sum_{k=1}^N -q_k\log\frac{\exp(s(V_i,T_{y_k}))}{\sum_{{y_a}=1}^N \exp(s(V_i,T_{y_a}))}
\end{equation}

Ultimately, the losses used in our second training stage are summarized as follows:
\begin{equation}\label{eq:stage2loss}
\mathcal L_{stage2} =\mathcal{L}_{id} + \mathcal{L}_{tri} +\mathcal{L}_{i2tce} 
\end{equation}

The whole training process of the proposed CLIP-ReID, including both the first and second stages, is summarized in \cref{alg:algorithm}. We use the learnable prompts to mine and store the hidden states of the pre-trained image encoder and text encoder, allowing CLIP to retain its own advantages. During the second stage, these prompts can regularize the image encoder and thus increase its generalization ability.

\subsubsection{SIE and OLP.} To make the model aware of the camera or viewpoint, we use Side Information Embeddings (SIE) \cite{transreid} to introduce relevant information. Unlike TransReID, we only add camera information to the [CLS] token, rather than all tokens, to avoid disturbing image details. Overlapping Patches (OLP) can further enhance the model with increased computational resources, which is realized simply by changing the stride in the token embedding.

\section{Experiments}
\subsection{Datasets and Evaluation Protocols}
We evaluate our method on four person re-identification datasets, including MSMT17 \cite{MSMT17}, Market-1501 \cite{market1501}, DukeMTMC-reID \cite{Duke}, Occluded-Duke \cite{Occduke}, and two vehicle ReID datasets, VeRi-776 \cite{VeRi} and VehicleID \cite{VehicleID}. The details of these datasets are summarized in \cref{tab:dataset}. Following common practices, we adapt the cumulative matching characteristics (CMC) at Rank-1 (R1) and the mean average precision (mAP) to evaluate the performance.

\subsection{Implementations}
\subsubsection{Models.} We adopt the visual %domain 
encoder $\mathcal{I}(\cdot)$ and the text %domain 
encoder  $\mathcal{T}(\cdot)$ from CLIP as the backbone for our image and text feature extractor. CLIP provides two alternatives $\mathcal{I}(\cdot)$, namely a transformer and a CNN with a global attention pooling layer. For the transformer, we choose the ViT-B/16, which contains 12 transformer layers with the hidden size of 768 dimensions. To match the output of the $\mathcal{T}(\cdot)$, the dimension of the image feature vector is reduced from 768 to 512 by a linear layer. For the CNN, we choose ResNet-50, where the last stride changes from 2 to 1, resulting in a larger feature map to preserve spatial information. The global attention pooling layer after ResNet-50 reduces the dimension of the embedding vectors from 2048 to 1024, matching the dimensions of the text features converted from 512 to 1024.

\begin{table}[]
\centering
\begin{tabular}{l|c c c c}
    \hline
    Dataset  & Image & ID & Cam + View\\
    \hline
    MSMT17 & 126,441 & 4,101 & 15\\
    Market-1501 & 32,668 & 1,501 & 6\\
    DukeMTMC-reID & 36,411 & 1,404 & 8 \\
    Occluded-Duke & 35,489 & 1,404 & 8 \\
    VeRi-776 & 49,357 & 776 & 28 \\
    VehicleID & 221,763 & 26,267 & - \\
    \hline
\end{tabular}
\caption{Statistics of datasets used in the paper.}
\label{tab:dataset}
\end{table}

\subsubsection{Training details.} In the first training stage, we use the Adam optimizer for both the CNN-based and the ViT-based models, with a learning rate initialized at $3.5 \times 10^{-4}$ and decayed by a cosine schedule. At this stage, the batch size is set to 64 without using any augmentation methods. Only the learnable text tokens $\rm[X]_1[X]_2[X]_3...[X]_M$ are optimizable. In the second training stage (same as our baseline), Adam optimizer is also used to train the image encoder. Each mini-batch consists of $\rm B = P \times K$ images, where $\rm P$ is the number of randomly selected identities, and $\rm K$ is samples per identity. We take $\rm P = 16$ and $\rm K= 4$. Each image is augmented by random horizontal flipping, padding, cropping and erasing \cite{randomerasing}. For the CNN-based model, we spend 10 epochs linearly increasing the learning rate from $3.5\times10^{-6}$ to $3.5\times10^{-4}$, and then the learning rate is decayed by 0.1 at the 40th and 70th epochs. For the ViT-based model, we warm up the model for 10 epochs with a linearly growing learning rate from $5 \times 10^{-7}$ to $5 \times 10^{-6}$. Then, it is decreased by a factor of 0.1 at the 30th and 50th epochs. We train the CNN-based model for 120 epochs while the ViT-based model for 60 epochs. For the CNN-based model, we use $\mathcal{L}_{tri}$ and $\mathcal{L}_{id}$ pre and post the global attention pooling layer, and $\alpha$ is set to 0.3. Similarly, we use them pre and post the linear layer after the transformer. Note that we also employ $\mathcal{L}_{tri}$  after the 11th transformer layer of ViT-B/16 and the 3rd residual layer of ResNet-50.

\begin{table*}[t]
\centering
\begin{tabular}{c|l|l|llllllll}
\hline
\multicolumn{1}{c|}{\multirow{2}{*}{Backbone}} & \multicolumn{1}{c|}{\multirow{2}{*}{Methods}}  & \multicolumn{1}{c|}{\multirow{2}{*}{References}} & \multicolumn{2}{c}{MSMT17} & \multicolumn{2}{c}{Market-1501} & \multicolumn{2}{c}{DukeMTMC} & \multicolumn{2}{c}{Occluded-Duke} \\
\multicolumn{1}{c|}{} & \multicolumn{1}{c|}{} & \multicolumn{1}{c|}{} & \multicolumn{1}{c}{mAP} & \multicolumn{1}{c}{R1} & \multicolumn{1}{c}{mAP} & \multicolumn{1}{c}{R1} & \multicolumn{1}{c}{mAP} & \multicolumn{1}{c}{R1} & \multicolumn{1}{c}{mAP} & \multicolumn{1}{c}{R1} \\ \hline
\multirow{12}{*}{CNN} & PCB* & ECCV \shortcite{PCB} & - & - & 81.6 & 93.8 & 69.2 & 83.3 & - & - \\
 & MGN* & MM \shortcite{MGN} & - & - & 86.9 & 95.7 & 78.4 & 88.7 & - & - \\
 & OSNeT& ICCV \shortcite{OSNeT} & 52.9 & 78.7 & 84.9 & 94.8 & 73.5 & 88.6 & - & - \\
 & ABD-Net*& ICCV \shortcite{ABD-Net}& 60.8 & 82.3 & 88.3 & 95.6 & 78.6 & 89.0 & - & - \\
 & Auto-ReID* & ICCV \shortcite{AutoREID} & 52.5 & 78.2 & 85.1& 94.5 & - & - & - & - \\
 & HOReID & CVPR \shortcite{HOReID}& - & - & 84.9 & 94.2 & 75.6 & 86.9 & 43.8 & 55.1 \\
 & ISP & ECCV \shortcite{ISP} & - & - & 88.6 & 95.3 & 80.0 & 89.6 & 52.3 & 62.8 \\
 & SAN & AAAI \shortcite{SAN-person} & 55.7 & 79.2 & 88.0 & \textbf{96.1} & 75.5 & 87.9 & - & - \\
 & OfM & AAAI \shortcite{OFM} & 54.7 & 78.4 & 87.9 & 94.9 & 78.6 & 89.0 & - & - \\ 
 & CDNet & CVPR \shortcite{CDNet} & 54.7 & 78.9 & 86.0 & 95.1 & 76.8 & 88.6 & - & - \\
 & PAT & CVPR \shortcite{PAT}& - & - & 88.0 & 95.4 & 78.2 & 88.8 & \textbf{53.6} & \textbf{64.5} \\
 & CAL* & ICCV \shortcite{CAL}& 56.2 & 79.5 & 87.0 & 94.5 & 76.4 & 87.2 & - & - \\
 & CBDB-Net* & TCSVT \shortcite{CBDB-Net} & - & - & 85.0 & 94.4 & 74.3 & 87.7 & 38.9 & 50.9 \\ 
 & ALDER* & TIP \shortcite{ALDER}& 59.1 & 82.5 & 88.9 & 95.6 & 78.9 & 89.9 & - & - \\
 & LTReID* & TMM \shortcite{LTReID}& 58.6 & 81.0 & 89.0 & 95.9 & 80.4 & \textbf{90.5} & - & - \\
 & DRL-Net & TMM \shortcite{DRL-Net} & 55.3 & 78.4 & 86.9 & 94.7 & 76.6 & 88.1 & 50.8 & 65.0 \\ 
 \cline{2-11} 
 & baseline &  & 60.7 & 82.1 & 88.1 & 94.7 & 79.3 & 88.6 & 47.4 & 54.2 \\
 & CLIP-ReID &  & \textbf{63.0} & \textbf{84.4} & \textbf{89.8} & 95.7 & \textbf{80.7} & 90.0 & 53.5 & 61.0 \\ \hline
\multirow{7}{*}{ViT} 
 & AAformer* & arxiv \shortcite {AAformer} &  63.2 & 83.6 & 87.7 & 95.4 & 80.0 & 90.1 & 58.2 & 67.0 \\
 & TransReID+SIE+OLP & ICCV \shortcite {transreid} &  67.4 & 85.3 & 88.9 & 95.2 & 82.0 & 90.7 & 59.2 & 66.4 \\
 & TransReID+SIE+OLP* & & 69.4 & 86.2 & 89.5 & 95.2 & 82.6 & 90.7 & - & - \\
 & DCAL & CVPR \shortcite{DCAL} & 64.0 & 83.1 & 87.5 & 94.7 & 80.1 & 89.0 & - & - \\
 \cline{2-11}
 & baseline & & 66.1 & 84.4 & 86.4 & 93.3 & 80.0 & 88.8 & 53.5 & 60.8 \\
 & CLIP-ReID & & \textbf{73.4} & \textbf{88.7} & \textbf{89.6} & \textbf{95.5} & 82.5 & 90.0 &  \textbf{59.5} &  \textbf{67.1} \\
 & CLIP-ReID+SIE+OLP & &\textbf{75.8} & \textbf{89.7} & \textbf{90.5} & \textbf{95.4} & \textbf{83.1} &  \textbf{90.8} &  \textbf{60.3} &  \textbf{67.2} \\ \hline
\end{tabular}
\caption{Comparison with state-of-the-art CNN- and ViT- based methods on person ReID datasets. DukeMTMC denotes the DukeMTMC-reID benchmark. The superscript star* means that the input image is resized to a resolution larger than 256x128.}
\label{tab:personmethod}
\end{table*}

\begin{table}[]
\centering
\begin{tabular}{c|l|cccc}
\hline
\multicolumn{1}{c|}{\multirow{2}{*}{\makecell[c]{Back\\-bone}}} & \multicolumn{1}{c|}{\multirow{2}{*}{Methods}} & \multicolumn{2}{c}{VeRi-776} & \multicolumn{2}{c}{VehicleID} \\
\multicolumn{1}{c|}{}& \multicolumn{1}{c|}{} & mAP & R1 & R1 & R5 \\ \hline

\multirow{12}{*}{CNN} & PRN \shortcite{PRN} & 74.3 & 94.3 & 78.4 & 92.3 \\
%  & VANet \shortcite{VANet} & 66.3 & 89.8 & 83.3 & 96.0 \\
 & PGAN \shortcite{PGAN} & 79.3 & 96.5 & 77.8 & 92.1 \\ 
 & SAN  \shortcite{SAN} & 72.5 & 93.3 & 79.7 & 94.3 \\ 
 & UMTS \shortcite{UMTS} & 75.9 & 95.8 & 80.9 & - \\
 & SPAN \shortcite{SPAN} & 68.9 & 94.0 & - & - \\
 & PVEN \shortcite{PVEN} & 79.5& 95.6 & 84.7 & 97.0 \\
 & SAVER \shortcite{SAVER} & 79.6 & 96.4 & 79.9 & 95.2 \\
 & CFVMNet \shortcite{CFVMNet} & 77.1 & 95.3 & 81.4 & 94.1\\ 
%  & GLAMOR \shortcite{GLAMOR} & 80.3 & 96.5 & 78.6 & 93.6\\ % 
%  & MDIM  \shortcite{MDIM} & 79.8 & 95.0 & - & - \\ 
 & CAL  \shortcite{CAL} & 74.3 & 95.4 & 82.5 & 94.7 \\ 
 & EIA-Net \shortcite{EIA-Net} & 79.3 & 95.7 & 84.1 & 96.5 \\
 & FIDI \shortcite{FIDI} & 77.6 & 95.7 & 78.5 & 91.9 \\
\cline{2-6} 
 & baseline & 79.3 & 95.7 & 84.4 & 96.6 \\
 & CLIP-ReID & \textbf{80.3} & \textbf{96.8} & \textbf{85.2} & \textbf{97.1} \\ \hline
\multirow{6}{*}{ViT} 
 & TransReID \shortcite{transreid} & 80.6 & 96.9 & 83.6 & 97.1 \\ 
 & TransReID! & 82.0 & 97.1 & 85.2 & 97.5 \\ 
 & DCAL \shortcite{DCAL} & 80.2 & 96.9 & - & - \\
 \cline{2-6} 
 & baseline  & 79.3 & 95.7 & 84.2 & 96.6 \\
 & CLIP-ReID  & \textbf{83.3} & \textbf{97.4} & \textbf{85.3} & \textbf{97.6} \\
 & CLIP-ReID! & \textbf{84.5} & \textbf{97.3} & \textbf{85.5} & 97.2 \\ \hline
\end{tabular}
\caption{Comparison with state-of-the-art CNN- and ViT- based methods on vehicle ReID datasets. Only the small subset of VehicleID is used in this paper. ! indicates that the method further uses SIE and OLP on VeRi-776 and OLP on VehicleID.}
\label{tab:carmethod}
\end{table}

\subsection{Comparison with State-of-the-Art Methods}
We compare our method with the state-of-the-art methods on three widely used person ReID benchmarks, one occluded ReID benchmark in \cref{tab:personmethod},  and two vehicle ReID benchmarks in \cref{tab:carmethod}. Despite being simple, CLIP-ReID achieves a strikingly good result. Note that all data listed here are without re-ranking.
\subsubsection{Person ReID.}
For both CNN-based and ViT-based methods, CLIP-ReID outperforms previous methods by a large margin on the most challenging dataset, MSMT17. Our method achieves $63.0\%$ mAP and $84.4\%$ R1 on the CNN-based backbone, and $73.4\%$ mAP and $88.7\%$ R1 ($6.0\%$ and $3.4\%$ higher than Transreid+SIE+OLP) on the ViT-based backbone using only the CLIP-ReID method, in further use of SIE and OLP we can improve mAP and R1 to $75.8\%$ and $89.7\%$. On other smaller or occluded datasets, such as Market1501, DukeMTMC-reID, and Occluded-Duke, we also increase the mAP with the ViT-based backbone by $1.0\%$, $0.5\%$ and $1.1\%$, respectively.

\subsubsection{Vehicle ReID.}
Our method achieves competitive performance compared to the prior CNN-based and ViT-based methods. %Particularly, 
With the ViT-based backbone, CLIP-ReID reaches $85.3\%$ mAP and $97.6\%$ R1 on VehicleID, while CLIP-ReID! reaches $84.5\%$ mAP and $97.3\%$ R1 on VeRi-776.

\subsection{Ablation Studies and Analysis}
We conduct comprehensive ablation studies on MSMT17 dataset to analyze the influences and sensitivity of some major parameters. 
\subsubsection{Baseline comparison.}
Many CNN-based works are based on the strong baseline proposed by BoT \cite{strongbaseline}. For ViT-based methods, TransReID's baseline is widely adopted, while AAformer also proposes a baseline. Although slightly different, both of them are pre-trained on ImageNet, which is different from ours. As shown in \cref{tab:baseline}, due to the effectiveness of CLIP pre-training, our baseline achieves superior performance compared to other baselines.

\begin{table}[]
\centering
\begin{tabular}{c|l|cc}
\hline
Backbones & \multicolumn{1}{c|}{Methods} & mAP & Rank-1 \\ \hline
\multirow{2}{*}{CNN} & BoT & 51.3 & 75.3 \\
 & CLIP-ReID baseline & \textbf{60.7} & \textbf{82.1} \\ \hline
\multirow{3}{*}{ViT} & AAformer baseline & 58.5 & 79.4 \\
 & TransReID baseline & 61.0 & 81.8 \\
 & CLIP-ReID baseline & \textbf{66.1} & \textbf{84.4} \\ \hline
\end{tabular}
\caption{Comparison of baselines on the MSMT17 dataset.}
\label{tab:baseline} 
\end{table}

\subsubsection{Necessity of two-stage training.}
CLIP aligns embeddings from text and image domains, so it is important to exploit its text encoder. Since ReID has no specific text that distinguishes different IDs, we aim to provide this by pre-training a set of learnable text tokens. There are two ways to optimize them. One is one-stage training, in which we train the image encoder $\mathcal{I}(\cdot)$ while using contrastive loss to train the text tokens at the same time. The other is the two-stage that we propose, in which we tune the learnable text tokens in the first stage and use them to calculate the $\mathcal{L}_{i2tce}$ in the second stage. To verify which approach is more effective, we perform a comparison on MSMT17. As shown in \cref{tab:stage}, the one-stage training is less effective because, in the early stage of training, learnable text tokens cannot describe the image well but affects the optimization of $\mathcal{I}(\cdot)$.

\begin{table}[]
\centering
\begin{tabular}{l|l|cc}
\hline
Backbone & Methods & mAP & Rank-1 \\ \hline
\multirow{3}{*}{CNN} & baseline & 60.7 & 82.1 \\
 & one stage & 61.9 & 82.8 \\
 & two stage & \textbf{63.0} & \textbf{84.4} \\ \hline
\multirow{3}{*}{ViT} & baseline & 66.1 & 84.4 \\
 & one stage & 68.9 & 85.9 \\
 & two stage & \textbf{73.4} & \textbf{88.7} \\ \hline
\end{tabular}
\caption{Comparison between one- %stage 
and two-stage training.}
\label{tab:stage}
% \vspace{-0.2cm}
\end{table}

\subsubsection{Constraint from text encoder in the second stage. }
There are $\rm P$ different IDs in a batch, with $\rm K$ images per ID. When computing $\mathcal{L}_{i2tce}$, if we only consider text embeddings for the IDs within a batch, like $\mathcal{L}_{i2t}$, the number of participating IDs is much less than the total number of IDs as in $\mathcal{L}_{id}$. We extend it to all IDs in the training set, like $\mathcal{L}_{i2tce}$. From \cref{tab:loss}, we can conclude that comparing with all IDs in the training set is better than only comparing with the IDs of the current batch. Another conclusion is that $\mathcal{L}_{t2i}$ is not necessary in the second stage. Finally, we combine the $\mathcal{L}_{id}$, $\mathcal{L}_{tri}$, $\mathcal{L}_{i2tce}$ to form the total loss. For the ViT, the weights of the three loss terms are 0.25, 1, and 1, respectively, while they are 1, 1, and 1 for the CNN.

\begin{table}[h]
\centering
\begin{tabular}{c|c|c|cc}
    \hline
    $\mathcal{L}_{i2tce}$ & $\mathcal{L}_{i2t}$ & $\mathcal{L}_{t2i}$ & mAP & Rank-1\\
    \hline
    - & - & - & 66.1 & 84.4\\
    - & \checkmark & \checkmark  & 71.3 & 87.5\\
    - & \checkmark & -  & 71.7 & 87.6\\
    \checkmark & - &\checkmark  & 73.2 & 88.6\\
    \checkmark & - & - & \textbf{73.4} &  \textbf{88.7}\\
    \hline
\end{tabular}
\caption{Loss terms from text encoder in the second stage.}
\label{tab:loss}
\end{table}

\subsubsection{Number of learnable tokens M.}
To be consistent with CLIP, we set the text description to ``A photo of a $\rm[X]_1[X]_2[X]_3...[X]_M$ person/vehicle.". We conduct analysis on the parameter $\rm M$ and find that $\rm M=1$ results in not learning sufficient text description, but when $\rm M$ is added to 8, it is redundant and unhelpful. We finally choose $\rm M=4$, which gives the best result among different settings.

\subsubsection{SIE and OLP.}
In \cref{tab:sieoverlap}, we evaluate the effectiveness of SIE and OLP on MSMT17. Using SIE only for [CLS] tokens works better than adding it for all global tokens. It gains $1.1\%$ mAP improvement on MSMT17 when the model uses only SIE-cls and $1.2\%$ improvement using only OLP. When applied together, mAP and R1 raise $2.4\%$ and $1.0\%$, respectively.

\begin{table}[h]
\centering
\begin{tabular}{lll|cc}
\hline
SIE-all & SIE-cls & OLP & mAP & Rank-1 \\ \hline
- & - & - & 73.4 & 88.7 \\
\checkmark & - & - & 74.3 & 88.6 \\
- & \checkmark & - & 74.5 & 88.8 \\
- & - & \checkmark & 74.6 & 89.5 \\
- & \checkmark & \checkmark & \textbf{75.8} & \textbf{89.7} \\ \hline
\end{tabular}
\caption{The validations on SIE-cls and OLP in ViT-based image encoder.}
\label{tab:sieoverlap}
\end{table}

\subsubsection{Visualization of CLIP-ReID.}
Finally, we perform visualization experiments using the \cite{visua} method to show the focused areas of the model. Both TransReID's and our baselines focus on local areas, ignoring other details about the human body, while CLIP-ReID will focus on a more comprehensive area.

\begin{figure}[h]
\centering
\includegraphics[width=\linewidth,scale=1.00]{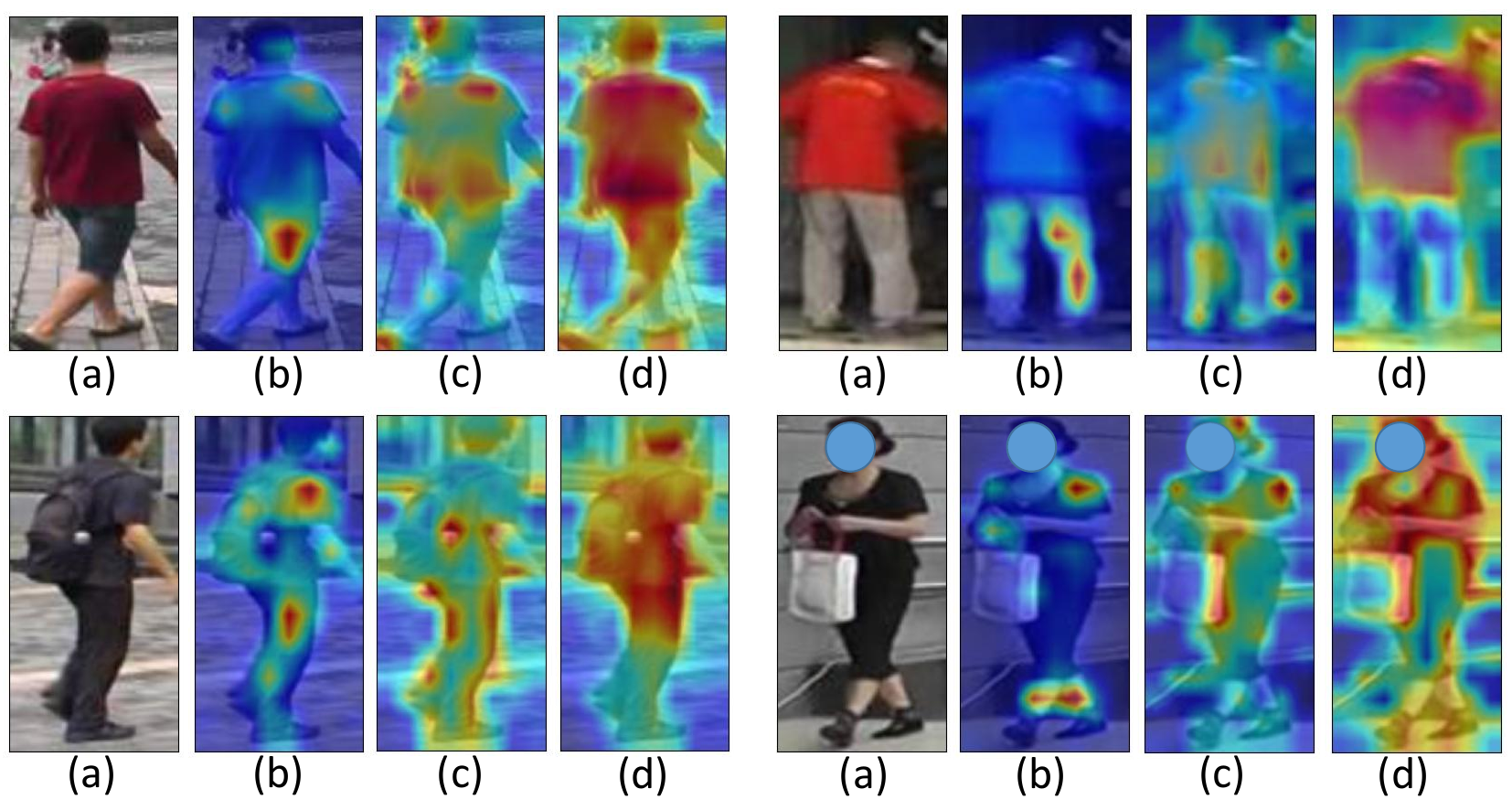} 
\caption{Visualization. % using the \cite{visua} method, 
(a) Input images, (b) TransReID baseline, (c) our baseline (d) CLIP-ReID.}
\label{fg:cam}
\end{figure}

\section{Conclusion}
This paper investigates the way to apply the vision-language pre-training model in image ReID. We find that fine-tuning the visual model initialized by the CLIP image encoder, either ResNet-50 or ViT-B/16, gives a good performance compared to other baselines. To fully utilize the cross-modal description ability in the pre-trained model, we propose CLIP-ReID with a two-stage training strategy, in which the learnable text tokens shared within each ID are incorporated and augmented to describe different instances. In the first stage, only these tokens get optimized, forming ambiguous text descriptions. In the second stage, these tokens and text encoder together provide constraints for optimizing the parameters in the image encoder. We validate CLIP-ReID on several datasets of persons and vehicles, and the results demonstrate the effectiveness of text descriptions and the superiority of our model. 

\section{Acknowledgments}
This work is supported by the Science and Technology Commission of Shanghai Municipality under Grant No. 22511105800, 19511120800 and 22DZ2229004.

% \bigskip
% \noindent Thank you for reading these instructions carefully. We look forward to receiving your electronic files!

% \bibliography{aaai23}

\begin{thebibliography}{65}
\providecommand{\natexlab}[1]{#1}

\bibitem[{Baldrati et~al.(2022)Baldrati, Bertini, Uricchio, and
  Del~Bimbo}]{CLIP4CirDemo}
Baldrati, A.; Bertini, M.; Uricchio, T.; and Del~Bimbo, A. 2022.
\newblock Effective Conditioned and Composed Image Retrieval Combining
  CLIP-Based Features.
\newblock In \emph{Proceedings of the IEEE/CVF CVPR}, 21466--21474.

\bibitem[{Chefer, Gur, and Wolf(2021)}]{visua}
Chefer, H.; Gur, S.; and Wolf, L. 2021.
\newblock Transformer interpretability beyond attention visualization.
\newblock In \emph{Proceedings of the IEEE/CVF CVPR}, 782--791.

\bibitem[{Chen et~al.(2019)Chen, Ding, Xie, Yuan, Chen, Yang, Ren, and
  Wang}]{ABD-Net}
Chen, T.; Ding, S.; Xie, J.; Yuan, Y.; Chen, W.; Yang, Y.; Ren, Z.; and Wang,
  Z. 2019.
\newblock Abd-net: Attentive but diverse person re-identification.
\newblock In \emph{Proceedings of the IEEE/CVF international conference on
  computer vision}, 8351--8361.

\bibitem[{Chen et~al.(2020)Chen, Liu, Wu, and Chien}]{SPAN}
Chen, T.-S.; Liu, C.-T.; Wu, C.-W.; and Chien, S.-Y. 2020.
\newblock Orientation-aware vehicle re-identification with semantics-guided
  part attention network.
\newblock In \emph{European conference on computer vision}, 330--346. Springer.

\bibitem[{Dai et~al.(2019)Dai, Chen, Gu, Zhu, and Tan}]{dai2019batch}
Dai, Z.; Chen, M.; Gu, X.; Zhu, S.; and Tan, P. 2019.
\newblock Batch dropblock network for person re-identification and beyond.
\newblock In \emph{Proceedings of the IEEE/CVF international conference on
  computer vision}, 3691--3701.

\bibitem[{Das, Chakraborty, and Roy-Chowdhury(2014)}]{das2014consistent}
Das, A.; Chakraborty, A.; and Roy-Chowdhury, A.~K. 2014.
\newblock Consistent re-identification in a camera network.
\newblock In \emph{European conference on computer vision}, 330--345. Springer.

\bibitem[{Dosovitskiy et~al.(2020)Dosovitskiy, Beyer, Kolesnikov, Weissenborn,
  Zhai, Unterthiner, Dehghani, Minderer, Heigold, Gelly et~al.}]{ViT}
Dosovitskiy, A.; Beyer, L.; Kolesnikov, A.; Weissenborn, D.; Zhai, X.;
  Unterthiner, T.; Dehghani, M.; Minderer, M.; Heigold, G.; Gelly, S.; et~al.
  2020.
\newblock An image is worth 16x16 words: Transformers for image recognition at
  scale.
\newblock \emph{arXiv preprint arXiv:2010.11929}.

\bibitem[{Gao et~al.(2021)Gao, Geng, Zhang, Ma, Fang, Zhang, Li, and
  Qiao}]{CLIP-Adapter}
Gao, P.; Geng, S.; Zhang, R.; Ma, T.; Fang, R.; Zhang, Y.; Li, H.; and Qiao, Y.
  2021.
\newblock Clip-adapter: Better vision-language models with feature adapters.
\newblock \emph{arXiv preprint arXiv:2110.04544}.

\bibitem[{Gu et~al.(2021)Gu, Lin, Kuo, and Cui}]{gu2021open}
Gu, X.; Lin, T.-Y.; Kuo, W.; and Cui, Y. 2021.
\newblock Open-vocabulary object detection via vision and language knowledge
  distillation.
\newblock \emph{arXiv preprint arXiv:2104.13921}.

\bibitem[{He et~al.(2019)He, Li, Zhao, and Tian}]{PRN}
He, B.; Li, J.; Zhao, Y.; and Tian, Y. 2019.
\newblock Part-regularized near-duplicate vehicle re-identification.
\newblock In \emph{Proceedings of the IEEE/CVF CVPR}, 3997--4005.

\bibitem[{He et~al.(2016)He, Zhang, Ren, and Sun}]{ResNet}
He, K.; Zhang, X.; Ren, S.; and Sun, J. 2016.
\newblock Deep residual learning for image recognition.
\newblock In \emph{Proceedings of the IEEE Conference on Computer Vision and
  Pattern Recognition (CVPR)}, 770--778.

\bibitem[{He et~al.(2021)He, Luo, Wang, Wang, Li, and Jiang}]{transreid}
He, S.; Luo, H.; Wang, P.; Wang, F.; Li, H.; and Jiang, W. 2021.
\newblock Transreid: Transformer-based object re-identification.
\newblock In \emph{Proceedings of the IEEE/CVF international conference on
  computer vision}, 15013--15022.

\bibitem[{Hermans, Beyer, and Leibe(2017)}]{Triplet}
Hermans, A.; Beyer, L.; and Leibe, B. 2017.
\newblock In defense of the triplet loss for person re-identification.
\newblock \emph{arXiv preprint arXiv:1703.07737}.

\bibitem[{Jia et~al.(2021)Jia, Yang, Xia, Chen, Parekh, Pham, Le, Sung, Li, and
  Duerig}]{ALIGN}
Jia, C.; Yang, Y.; Xia, Y.; Chen, Y.-T.; Parekh, Z.; Pham, H.; Le, Q.; Sung,
  Y.-H.; Li, Z.; and Duerig, T. 2021.
\newblock Scaling up visual and vision-language representation learning with
  noisy text supervision.
\newblock In \emph{International Conference on Machine Learning}, 4904--4916.
  PMLR.

\bibitem[{Jia et~al.(2022)Jia, Cheng, Lu, and Zhang}]{DRL-Net}
Jia, M.; Cheng, X.; Lu, S.; and Zhang, J. 2022.
\newblock Learning disentangled representation implicitly via transformer for
  occluded person re-identification.
\newblock \emph{IEEE Transactions on Multimedia}.

\bibitem[{Jin et~al.(2020{\natexlab{a}})Jin, Lan, Zeng, and Chen}]{UMTS}
Jin, X.; Lan, C.; Zeng, W.; and Chen, Z. 2020{\natexlab{a}}.
\newblock Uncertainty-aware multi-shot knowledge distillation for image-based
  object re-identification.
\newblock In \emph{Proceedings of the AAAI Conference on Artificial
  Intelligence}, volume~34, 11165--11172.

\bibitem[{Jin et~al.(2020{\natexlab{b}})Jin, Lan, Zeng, Wei, and
  Chen}]{SAN-person}
Jin, X.; Lan, C.; Zeng, W.; Wei, G.; and Chen, Z. 2020{\natexlab{b}}.
\newblock Semantics-aligned representation learning for person
  re-identification.
\newblock In \emph{Proceedings of the AAAI Conference on Artificial
  Intelligence}, volume~34, 11173--11180.

\bibitem[{Khorramshahi et~al.(2020)Khorramshahi, Peri, Chen, and
  Chellappa}]{SAVER}
Khorramshahi, P.; Peri, N.; Chen, J.-c.; and Chellappa, R. 2020.
\newblock The devil is in the details: Self-supervised attention for vehicle
  re-identification.
\newblock In \emph{European Conference on Computer Vision}, 369--386. Springer.

\bibitem[{Kim, Son, and Kim(2021)}]{ViLT}
Kim, W.; Son, B.; and Kim, I. 2021.
\newblock Vilt: Vision-and-language transformer without convolution or region
  supervision.
\newblock In \emph{International Conference on Machine Learning}, 5583--5594.
  PMLR.

\bibitem[{Koestinger et~al.(2012)Koestinger, Hirzer, Wohlhart, Roth, and
  Bischof}]{koestinger2012large}
Koestinger, M.; Hirzer, M.; Wohlhart, P.; Roth, P.~M.; and Bischof, H. 2012.
\newblock Large scale metric learning from equivalence constraints.
\newblock In \emph{2012 IEEE CVPR}, 2288--2295. IEEE.

\bibitem[{Li, Wu, and Zheng(2021)}]{CDNet}
Li, H.; Wu, G.; and Zheng, W.-S. 2021.
\newblock Combined depth space based architecture search for person
  re-identification.
\newblock In \emph{Proceedings of the IEEE/CVF CVPR}, 6729--6738.

\bibitem[{Li et~al.(2022)Li, Li, Xiong, and Hoi}]{li2022blip}
Li, J.; Li, D.; Xiong, C.; and Hoi, S. 2022.
\newblock Blip: Bootstrapping language-image pre-training for unified
  vision-language understanding and generation.
\newblock \emph{arXiv preprint arXiv:2201.12086}.

\bibitem[{Li et~al.(2021{\natexlab{a}})Li, Selvaraju, Gotmare, Joty, Xiong, and
  Hoi}]{ALBEF}
Li, J.; Selvaraju, R.; Gotmare, A.; Joty, S.; Xiong, C.; and Hoi, S. C.~H.
  2021{\natexlab{a}}.
\newblock Align before fuse: Vision and language representation learning with
  momentum distillation.
\newblock \emph{Advances in neural information processing systems}, 34:
  9694--9705.

\bibitem[{Li et~al.(2021{\natexlab{b}})Li, He, Zhang, Liu, Zhang, and Wu}]{PAT}
Li, Y.; He, J.; Zhang, T.; Liu, X.; Zhang, Y.; and Wu, F. 2021{\natexlab{b}}.
\newblock Diverse part discovery: Occluded person re-identification with
  part-aware transformer.
\newblock In \emph{Proceedings of the IEEE/CVF CVPR}, 2898--2907.

\bibitem[{Liang et~al.(2018)Liang, Lang, Li, Zhao, Wang, and Feng}]{EIA-Net}
Liang, L.; Lang, C.; Li, Z.; Zhao, J.; Wang, T.; and Feng, S. 2018.
\newblock Seeing Crucial Parts: Vehicle Model Verification via A Discriminative
  Representation Model.
\newblock \emph{Journal of the ACM (JACM)}, 22.

\bibitem[{Liao et~al.(2015)Liao, Hu, Zhu, and Li}]{liao2015person}
Liao, S.; Hu, Y.; Zhu, X.; and Li, S.~Z. 2015.
\newblock Person re-identification by local maximal occurrence representation
  and metric learning.
\newblock In \emph{Proceedings of the IEEE CVPR}, 2197--2206.

\bibitem[{Liu et~al.(2016{\natexlab{a}})Liu, Tian, Yang, Pang, and
  Huang}]{VehicleID}
Liu, H.; Tian, Y.; Yang, Y.; Pang, L.; and Huang, T. 2016{\natexlab{a}}.
\newblock Deep relative distance learning: Tell the difference between similar
  vehicles.
\newblock In \emph{Proceedings of the IEEE CVPR}, 2167--2175.

\bibitem[{Liu et~al.(2016{\natexlab{b}})Liu, Liu, Ma, and Fu}]{VeRi}
Liu, X.; Liu, W.; Ma, H.; and Fu, H. 2016{\natexlab{b}}.
\newblock Large-scale vehicle re-identification in urban surveillance videos.
\newblock In \emph{2016 IEEE international conference on multimedia and expo
  (ICME)}, 1--6. IEEE.

\bibitem[{Luo et~al.(2019)Luo, Gu, Liao, Lai, and Jiang}]{strongbaseline}
Luo, H.; Gu, Y.; Liao, X.; Lai, S.; and Jiang, W. 2019.
\newblock Bag of tricks and a strong baseline for deep person
  re-identification.
\newblock In \emph{Proceedings of the IEEE/CVF CVPR workshops}, 0--0.

\bibitem[{Ma et~al.(2022)Ma, Zhao, Lin, Kale, Wang, Yu, Gu, Choudhary, and
  Xie}]{EL-CLIP}
Ma, H.; Zhao, H.; Lin, Z.; Kale, A.; Wang, Z.; Yu, T.; Gu, J.; Choudhary, S.;
  and Xie, X. 2022.
\newblock EI-CLIP: Entity-Aware Interventional Contrastive Learning for
  E-Commerce Cross-Modal Retrieval.
\newblock In \emph{Proceedings of the IEEE/CVF CVPR}, 18051--18061.

\bibitem[{Matsukawa et~al.(2016)Matsukawa, Okabe, Suzuki, and
  Sato}]{matsukawa2016hierarchical}
Matsukawa, T.; Okabe, T.; Suzuki, E.; and Sato, Y. 2016.
\newblock Hierarchical gaussian descriptor for person re-identification.
\newblock In \emph{Proceedings of the IEEE CVPR}, 1363--1372.

\bibitem[{Meng et~al.(2020)Meng, Li, Liu, Li, Yang, Zha, Gao, Wang, and
  Huang}]{PVEN}
Meng, D.; Li, L.; Liu, X.; Li, Y.; Yang, S.; Zha, Z.-J.; Gao, X.; Wang, S.; and
  Huang, Q. 2020.
\newblock Parsing-based view-aware embedding network for vehicle
  re-identification.
\newblock In \emph{Proceedings of the IEEE/CVF CVPR}, 7103--7112.

\bibitem[{Miao et~al.(2019)Miao, Wu, Liu, Ding, and Yang}]{Occduke}
Miao, J.; Wu, Y.; Liu, P.; Ding, Y.; and Yang, Y. 2019.
\newblock Pose-guided feature alignment for occluded person re-identification.
\newblock In \emph{Proceedings of the IEEE/CVF international conference on
  computer vision}, 542--551.

\bibitem[{Qian et~al.(2020)Qian, Jiang, Luo, and Yu}]{SAN}
Qian, J.; Jiang, W.; Luo, H.; and Yu, H. 2020.
\newblock Stripe-based and attribute-aware network: A two-branch deep model for
  vehicle re-identification.
\newblock \emph{Measurement Science and Technology}, 31(9): 095401.

\bibitem[{Quan et~al.(2019)Quan, Dong, Wu, Zhu, and Yang}]{AutoREID}
Quan, R.; Dong, X.; Wu, Y.; Zhu, L.; and Yang, Y. 2019.
\newblock Auto-reid: Searching for a part-aware convnet for person
  re-identification.
\newblock In \emph{Proceedings of the IEEE/CVF International Conference on
  Computer Vision}, 3750--3759.

\bibitem[{Radford et~al.(2021)Radford, Kim, Hallacy, Ramesh, Goh, Agarwal,
  Sastry, Askell, Mishkin, Clark et~al.}]{CLIP}
Radford, A.; Kim, J.~W.; Hallacy, C.; Ramesh, A.; Goh, G.; Agarwal, S.; Sastry,
  G.; Askell, A.; Mishkin, P.; Clark, J.; et~al. 2021.
\newblock Learning transferable visual models from natural language
  supervision.
\newblock In \emph{International Conference on Machine Learning}, 8748--8763.
  PMLR.

\bibitem[{Rao et~al.(2021)Rao, Chen, Lu, and Zhou}]{CAL}
Rao, Y.; Chen, G.; Lu, J.; and Zhou, J. 2021.
\newblock Counterfactual attention learning for fine-grained visual
  categorization and re-identification.
\newblock In \emph{Proceedings of the IEEE/CVF International Conference on
  Computer Vision}, 1025--1034.

\bibitem[{Rao et~al.(2022)Rao, Zhao, Chen, Tang, Zhu, Huang, Zhou, and
  Lu}]{rao2022denseclip}
Rao, Y.; Zhao, W.; Chen, G.; Tang, Y.; Zhu, Z.; Huang, G.; Zhou, J.; and Lu, J.
  2022.
\newblock Denseclip: Language-guided dense prediction with context-aware
  prompting.
\newblock In \emph{Proceedings of the IEEE/CVF CVPR}, 18082--18091.

\bibitem[{Ristani et~al.(2016)Ristani, Solera, Zou, Cucchiara, and
  Tomasi}]{Duke}
Ristani, E.; Solera, F.; Zou, R.; Cucchiara, R.; and Tomasi, C. 2016.
\newblock Performance measures and a data set for multi-target, multi-camera
  tracking.
\newblock In \emph{European conference on computer vision}, 17--35. Springer.

\bibitem[{Sennrich, Haddow, and Birch(2015)}]{BPE}
Sennrich, R.; Haddow, B.; and Birch, A. 2015.
\newblock Neural machine translation of rare words with subword units.
\newblock \emph{arXiv preprint arXiv:1508.07909}.

\bibitem[{Sun et~al.(2018)Sun, Zheng, Yang, Tian, and Wang}]{PCB}
Sun, Y.; Zheng, L.; Yang, Y.; Tian, Q.; and Wang, S. 2018.
\newblock Beyond part models: Person retrieval with refined part pooling (and a
  strong convolutional baseline).
\newblock In \emph{Proceedings of the European conference on computer vision
  (ECCV)}, 480--496.

\bibitem[{Sun et~al.(2020)Sun, Nie, Xi, and Yin}]{CFVMNet}
Sun, Z.; Nie, X.; Xi, X.; and Yin, Y. 2020.
\newblock Cfvmnet: A multi-branch network for vehicle re-identification based
  on common field of view.
\newblock In \emph{Proceedings of the 28th ACM international conference on
  multimedia}, 3523--3531.

\bibitem[{Tan et~al.(2021)Tan, Liu, Bian, Wang, and Yin}]{CBDB-Net}
Tan, H.; Liu, X.; Bian, Y.; Wang, H.; and Yin, B. 2021.
\newblock Incomplete descriptor mining with elastic loss for person
  re-identification.
\newblock \emph{IEEE Transactions on Circuits and Systems for Video
  Technology}, 32(1): 160--171.

\bibitem[{Van~der Maaten and Hinton(2008)}]{t-sne}
Van~der Maaten, L.; and Hinton, G. 2008.
\newblock Visualizing data using t-SNE.
\newblock \emph{Journal of machine learning research}, 9(11).

\bibitem[{Wang et~al.(2020)Wang, Yang, Liu, Wang, Yang, Wang, Yu, Zhou, and
  Sun}]{HOReID}
Wang, G.; Yang, S.; Liu, H.; Wang, Z.; Yang, Y.; Wang, S.; Yu, G.; Zhou, E.;
  and Sun, J. 2020.
\newblock High-order information matters: Learning relation and topology for
  occluded person re-identification.
\newblock In \emph{Proceedings of the IEEE/CVF CVPR}, 6449--6458.

\bibitem[{Wang et~al.(2018)Wang, Yuan, Chen, Li, and Zhou}]{MGN}
Wang, G.; Yuan, Y.; Chen, X.; Li, J.; and Zhou, X. 2018.
\newblock Learning discriminative features with multiple granularities for
  person re-identification.
\newblock In \emph{Proceedings of the 26th ACM international conference on
  Multimedia}, 274--282.

\bibitem[{Wang et~al.(2022)Wang, Zhao, Su, and Meng}]{LTReID}
Wang, P.; Zhao, Z.; Su, F.; and Meng, H. 2022.
\newblock LTReID: Factorizable Feature Generation with Independent Components
  for Long-Tailed Person Re-Identification.
\newblock \emph{IEEE Transactions on Multimedia}.

\bibitem[{Wang et~al.(2021)Wang, Yu, Yu, Dai, Tsvetkov, and Cao}]{SIMVLM}
Wang, Z.; Yu, J.; Yu, A.~W.; Dai, Z.; Tsvetkov, Y.; and Cao, Y. 2021.
\newblock Simvlm: Simple visual language model pretraining with weak
  supervision.
\newblock \emph{arXiv preprint arXiv:2108.10904}.

\bibitem[{Wei et~al.(2018)Wei, Zhang, Gao, and Tian}]{MSMT17}
Wei, L.; Zhang, S.; Gao, W.; and Tian, Q. 2018.
\newblock Person transfer gan to bridge domain gap for person
  re-identification.
\newblock In \emph{Proceedings of the IEEE CVPR}, 79--88.

\bibitem[{Yan et~al.(2021)Yan, Pang, Bai, Liu, Ning, Gu, and Zhou}]{FIDI}
Yan, C.; Pang, G.; Bai, X.; Liu, C.; Ning, X.; Gu, L.; and Zhou, J. 2021.
\newblock Beyond triplet loss: person re-identification with fine-grained
  difference-aware pairwise loss.
\newblock \emph{IEEE Transactions on Multimedia}, 24: 1665--1677.

\bibitem[{Yi et~al.(2014)Yi, Lei, Liao, and Li}]{yi2014deep}
Yi, D.; Lei, Z.; Liao, S.; and Li, S.~Z. 2014.
\newblock Deep metric learning for person re-identification.
\newblock In \emph{2014 22nd international conference on pattern recognition},
  34--39. IEEE.

\bibitem[{Zhang et~al.(2021{\natexlab{a}})Zhang, Jiang, Cheng, Wu, Yu, Li, Guo,
  Zheng, Zheng, and Sun}]{OFM}
Zhang, E.; Jiang, X.; Cheng, H.; Wu, A.; Yu, F.; Li, K.; Guo, X.; Zheng, F.;
  Zheng, W.; and Sun, X. 2021{\natexlab{a}}.
\newblock One for More: Selecting Generalizable Samples for Generalizable ReID
  Model.
\newblock In \emph{Proceedings of the AAAI Conference on Artificial
  Intelligence}, volume~35, 3324--3332.

\bibitem[{Zhang et~al.(2021{\natexlab{b}})Zhang, Lai, Feng, and Xie}]{ALDER}
Zhang, Q.; Lai, J.; Feng, Z.; and Xie, X. 2021{\natexlab{b}}.
\newblock Seeing like a human: Asynchronous learning with dynamic progressive
  refinement for person re-identification.
\newblock \emph{IEEE Transactions on Image Processing}, 31: 352--365.

\bibitem[{Zhang et~al.(2019)Zhang, Zhang, Cao, Gong, You, and Shen}]{PGAN}
Zhang, X.; Zhang, R.; Cao, J.; Gong, D.; You, M.; and Shen, C. 2019.
\newblock Part-guided attention learning for vehicle re-identification.
\newblock \emph{arXiv preprint arXiv:1909.06023}, 2(8).

\bibitem[{Zhang et~al.(2021{\natexlab{c}})Zhang, He, Sun, and Li}]{hardmix}
Zhang, Y.; He, B.; Sun, L.; and Li, Q. 2021{\natexlab{c}}.
\newblock Progressive Multi-Stage Feature Mix for Person Re-Identification.
\newblock In \emph{ICASSP 2021-2021 IEEE International Conference on Acoustics,
  Speech and Signal Processing (ICASSP)}, 2765--2769. IEEE.

\bibitem[{Zhang et~al.(2020)Zhang, Lan, Zeng, Jin, and Chen}]{RGA}
Zhang, Z.; Lan, C.; Zeng, W.; Jin, X.; and Chen, Z. 2020.
\newblock Relation-aware global attention for person re-identification.
\newblock In \emph{Proceedings of the ieee/cvf CVPR}, 3186--3195.

\bibitem[{Zheng et~al.(2015)Zheng, Shen, Tian, Wang, Wang, and
  Tian}]{market1501}
Zheng, L.; Shen, L.; Tian, L.; Wang, S.; Wang, J.; and Tian, Q. 2015.
\newblock Scalable person re-identification: A benchmark.
\newblock In \emph{Proceedings of the IEEE international conference on computer
  vision}, 1116--1124.

\bibitem[{Zhong et~al.(2020)Zhong, Zheng, Kang, Li, and Yang}]{randomerasing}
Zhong, Z.; Zheng, L.; Kang, G.; Li, S.; and Yang, Y. 2020.
\newblock Random erasing data augmentation.
\newblock In \emph{Proceedings of the AAAI conference on artificial
  intelligence}, volume~34, 13001--13008.

\bibitem[{Zhou, Loy, and Dai(2021)}]{zhou2021denseclip}
Zhou, C.; Loy, C.~C.; and Dai, B. 2021.
\newblock Denseclip: Extract free dense labels from clip.
\newblock \emph{arXiv preprint arXiv:2112.01071}.

\bibitem[{Zhou et~al.(2021)Zhou, Yang, Loy, and Liu}]{CoOP}
Zhou, K.; Yang, J.; Loy, C.~C.; and Liu, Z. 2021.
\newblock Learning to Prompt for Vision-Language Models.

\bibitem[{Zhou et~al.(2022)Zhou, Yang, Loy, and Liu}]{CoCoOp}
Zhou, K.; Yang, J.; Loy, C.~C.; and Liu, Z. 2022.
\newblock Conditional prompt learning for vision-language models.
\newblock In \emph{Proceedings of the IEEE/CVF CVPR}, 16816--16825.

\bibitem[{Zhou et~al.(2019)Zhou, Yang, Cavallaro, and Xiang}]{OSNeT}
Zhou, K.; Yang, Y.; Cavallaro, A.; and Xiang, T. 2019.
\newblock Omni-scale feature learning for person re-identification.
\newblock In \emph{Proceedings of the IEEE/CVF International Conference on
  Computer Vision}, 3702--3712.

\bibitem[{Zhu et~al.(2022)Zhu, Ke, Li, Liu, Tian, and Shan}]{DCAL}
Zhu, H.; Ke, W.; Li, D.; Liu, J.; Tian, L.; and Shan, Y. 2022.
\newblock Dual Cross-Attention Learning for Fine-Grained Visual Categorization
  and Object Re-Identification.
\newblock In \emph{Proceedings of the IEEE/CVF CVPR}, 4692--4702.

\bibitem[{Zhu et~al.(2020)Zhu, Guo, Liu, Tang, and Wang}]{ISP}
Zhu, K.; Guo, H.; Liu, Z.; Tang, M.; and Wang, J. 2020.
\newblock Identity-guided human semantic parsing for person re-identification.
\newblock In \emph{European Conference on Computer Vision}, 346--363. Springer.

\bibitem[{Zhu et~al.(2021)Zhu, Guo, Zhang, Wang, Huang, Qiao, Liu, Wang, and
  Tang}]{AAformer}
Zhu, K.; Guo, H.; Zhang, S.; Wang, Y.; Huang, G.; Qiao, H.; Liu, J.; Wang, J.;
  and Tang, M. 2021.
\newblock Aaformer: Auto-aligned transformer for person re-identification.
\newblock \emph{arXiv preprint arXiv:2104.00921}.

\end{thebibliography}

% \fontsize{9.0pt}{10.0pt} \selectfont
\begin{small}

\end{small}

\appendix
\clearpage
\section{Supplementary Material}

\subsection{Alternative way for the first training stage. }
In order to improve the training efficiency in the first stage, we propose another method, which computes the $\mathcal{L}_{t2ice}$ based on the average image feature $V_{y_i}$ among all images with ID $y_i$. 
Since the image encoder keeps fixed, $V_{y_i}$ can be computed offline, and buffered in the memory. In this way, the first stage is completed in a short time at the expense of final performance dropping, shown in \cref{tab:onestageupdate}. 
\begin{equation}\label{eq:lt2ice}
\mathcal{L}_{t2ice}(y_i) = -\log\frac{\exp(s(V_{y_i},T_{y_i}))}{\sum_{{y_a}=1}^N \exp(s(V_{y_a},T_{y_i}))}
\end{equation}

\begin{table}[h]
\centering
\begin{tabular}{l|cccccc}
    \hline
    \multirow{2}{*}{Methods} & \multicolumn{3}{c}{MSMT17} & \multicolumn{3}{c}{Market-1501} \\
    & mAP & R1 & T & mAP & R1 & T\\ \hline
    average & 72.4 & 88.3 & 2.0 & 89.3 & 95.2 & 2.0\\
    instance & 73.4 & 88.6 & 37.5 & 89.6 & 95.5 & 15.5\\
    \hline
\end{tabular}
\caption{Comparison between two training methods in the first stage. T denotes the time of training in minutes. We implement ViT-based model on a single NVIDIA GeForce RTX 2080 Ti GPU. }
\label{tab:onestageupdate}
\end{table}

\subsection{Retrieval result visualization.}
We visualize the retrieval results on MSMT17, with the incorrectly identified samples highlighted in orange.

\begin{figure}[h]
\centering
\includegraphics[width=\linewidth,scale=1.00]{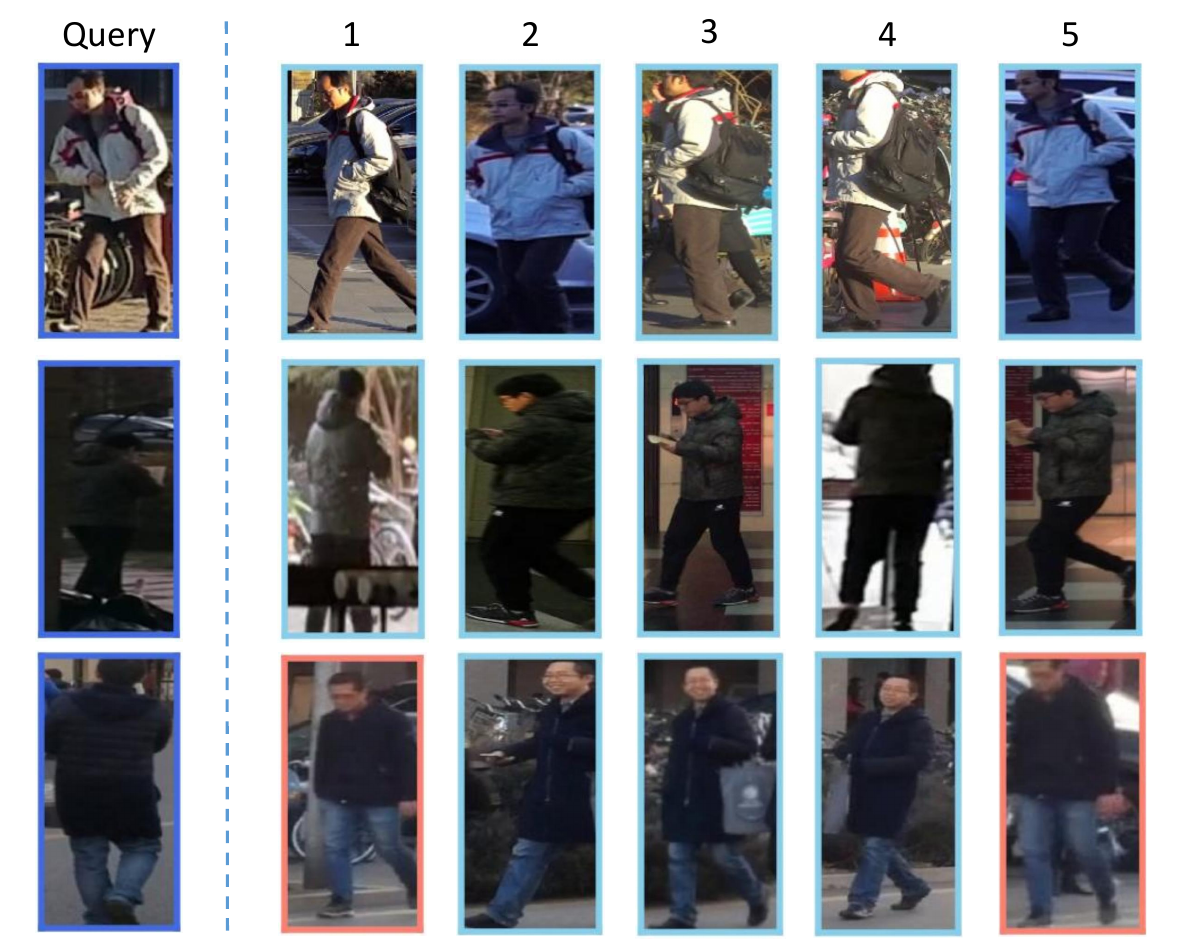}
\caption{Retrieval result visualization.}
\label{fg:inference}
\end{figure}

\subsection{The $\mathcal{L}_{tri}$ of the previous layer.}
We find that adding $\mathcal{L}_{tri}$ to the previous layer usually improves the model's ability to discriminate between different IDs, as shown in \cref{tab:wotri}. Therefore, we employ $\mathcal{L}_{tri}$ after the 11th transformer layer of ViT-B/16 and the 3rd residual layer of ResNet-50. To give a clear illustration of our loss constraint, we show them in the CNN version of CLIP-ReID in \cref{fg:inference}.

\begin{table}[h]
\centering
\begin{tabular}{c|l|cccc}
\hline
\multirow{2}{*}{\begin{tabular}[c]{@{}c@{}}Back-\\ bone\end{tabular}} & \multicolumn{1}{c|}{\multirow{2}{*}{Methods}} & \multicolumn{2}{c}{MSMT17} \\
 & \multicolumn{1}{c|}{} & mAP & R1\\ \hline
\multirow{4}{*}{CNN} 
 & baseline wo pre $\mathcal{L}_{tri}$ & 57.7 & 79.8 \\
 & baseline & \textbf{60.7} & \textbf{82.1} \\
 & CLIP-ReID wo pre $\mathcal{L}_{tri}$ & 59.5 & 82.1 \\ 
 & CLIP-ReID & \textbf{63.0} & \textbf{84.4}  \\\hline
\multirow{4}{*}{ViT} 
 & baseline wo pre $\mathcal{L}_{tri}$ & 65.9 & 84.2  \\
 & baseline & \textbf{66.0} & \textbf{84.4}  \\
 & CLIP-ReID wo pre $\mathcal{L}_{tri}$ & 73.1 & 88.6  \\ 
 & CLIP-ReID & \textbf{73.4} & \textbf{88.7}  \\\hline
\end{tabular}
\caption{Comparison of w/wo $\mathcal{L}_{tri}$ of the previous layer.}
\label{tab:wotri}
\end{table}

\subsection{Number of learnable tokens M.}
We conduct analysis on the number of learnable tokens $\rm M$ in \cref{tab:numM}. It shows that the performance is not sensitive to $\rm M$, and $\rm M=4$ gives the best results, as we present in the manuscript.

\begin{table}[h]
\centering
\begin{tabular}{l|cccc}
\hline
\multirow{2}{*}{$\rm M$} & \multicolumn{2}{c}{MSMT17} & \multicolumn{2}{c}{Market-1501} \\
 & mAP & R1 & mAP & R1 \\ \hline
1 & 72.7 & 88.4 & 89.5 & 94.9 \\
4 & \textbf{73.4} & \textbf{88.7} & \textbf{89.6} & \textbf{95.5} \\
8 & 73.4 & 88.6 & 89.5 & 95.1 \\ \hline
\end{tabular}
\caption{Performance analysis on parameter $\rm M$ for ViT-based.}
\label{tab:numM}
\end{table}

\subsection{Dimensions of inference features}
As shown in \cref{fg:inference}, we have three image features to use during inference, %namely pre\_img\_feature, img\_feature, post\_img\_feature, 
the results of different combinations in \cref{tab:inference}. We concatenate the img\_feature and post\_img\_feature as the final feature representation.

\subsection{Comparison with state-of-the-art methods on two vehicle datasets.}
We further intensively evaluate our proposed method on two vehicle ReID datasets, and the full metrics are shown in \cref{tab:vehicleall}. For the VehicleID dataset, the test sets are available in small, medium, and large versions, and CLIP-ReID achieves promising results in all the three settings.

\begin{figure*}[]
\centering
\includegraphics[width=\linewidth,scale=1.00]{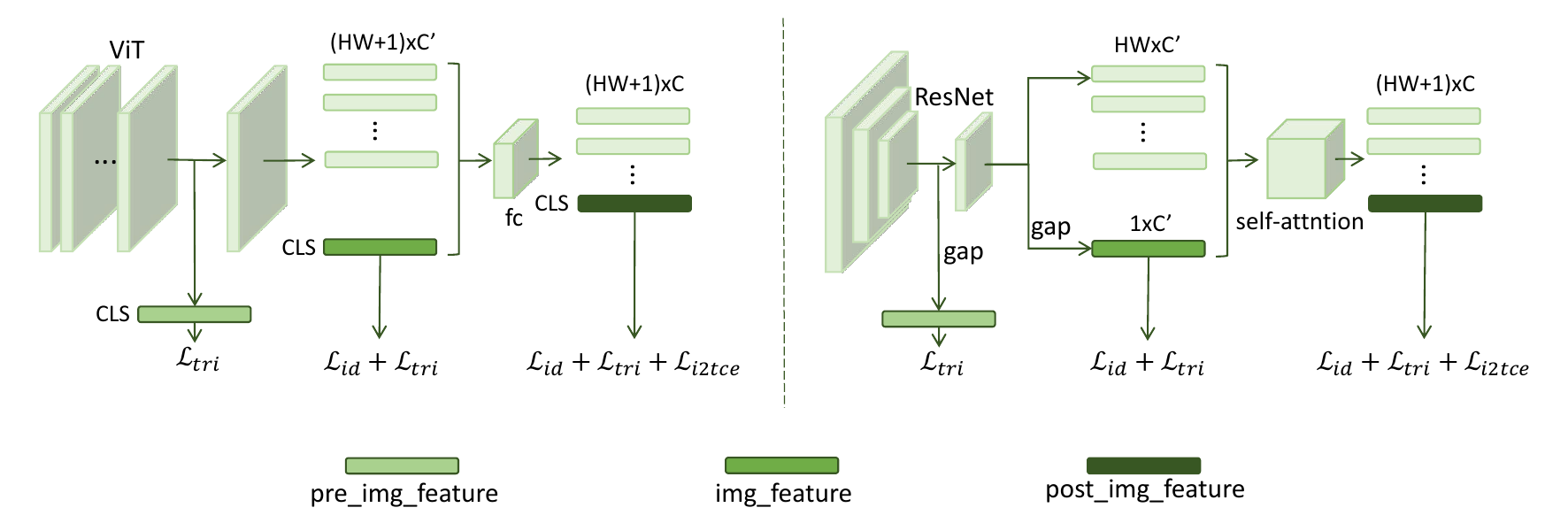}
\caption{Details of the CNN-based CLIP-ReID image encoder.}
\label{fg:inference}
\end{figure*}

\begin{table*}[]
\centering
\begin{tabular}{c|l|c|cccccccc}
\hline
\rm

% \multicolumn{1}{c|}{\multirow{2}{*}{Backbone}} & \multicolumn{1}{c|}{\multirow{2}{*}{Methods}}  & \multicolumn{1}{c|}{\multirow{2}{*}{References}} & \multicolumn{2}{c}{MSMT17} & \multicolumn{2}{c}{Market-1501} & \multicolumn{2}{c}{DukeMTMC} & \multicolumn{2}{c}{Occluded-Duke} \\

\multirow{2}{*}{Backbone} & \multirow{2}{*}{Inference features}  & \multirow{2}{*}{Dim} & \multicolumn{2}{c}{MSMT17} & \multicolumn{2}{c}{Market-1501} \\

% \multicolumn{1}{c}{\multirow{2}{*}{Backbone}} & \multicolumn{1}{c|}{\multirow{2}{*}{Inference features}}  & \multicolumn{1}{c|}{\multirow{2}{*}{Dim}} & \multicolumn{2}{c}{MSMT17} & \multicolumn{2}{c}{Market-1501} \\

\multicolumn{1}{c|}{} & \multicolumn{1}{c|}{} & \multicolumn{1}{c|}{} & \multicolumn{1}{c}{mAP} & \multicolumn{1}{c}{R1} & \multicolumn{1}{c}{mAP} & \multicolumn{1}{c}{R1} \\ \hline
\multirow{6}{*}{CNN} 
 & post\_img\_feature & 1024 & 61.3 & 84.0 & 88.6 & 95.2\\ 
 & pre\_img\_feature & 2048 & 48.3 & 68.8 & 83.1 & 91.9\\ 
 & img\_feature & 2048 & 57.6 & 80.5 & 88.6 & 95.2\\ 
 & \textbf{img\_feature + post\_img\_feature} & \textbf{3072} & 63.0 & 84.4 & 89.8 & 95.7 \\ 
 & img\_feature + pre\_img\_feature & 4096 & 57.0 & 78.7 & 88.5 & 94.9\\
 & pre\_img\_feature + img\_feature + post\_img\_feature & 5120 & 62.9 & 83.9 & 89.9 & 95.6\\ \hline
\multirow{6}{*}{ViT} 
 & post\_img\_feature & 512 & 72.3 & 88.2 & 89.0 & 94.9\\ 
 & pre\_img\_feature & 768 & 71.7 & 87.1 & 88.3 & 94.7\\
 & img\_feature & 768 & 73.4 & 88.7 & 89.6 & 95.4\\ 
 & \textbf{img\_feature + post\_img\_feature} & \textbf{1280} & 73.4 & 88.7 & 89.6 & 95.5 \\ 
 & img\_feature + pre\_img\_feature & 1536 & 73.6 & 88.6 & 89.7 & 95.5\\
 & pre\_img\_feature + img\_feature + post\_img\_feature & 2048 & 73.6 & 88.6 & 89.7 & 95.5\\ \hline
\end{tabular}
\caption{The validations on different inferece features.}
\label{tab:inference}
\end{table*}

\begin{table*}[]
\centering
\begin{tabular}{l|ccc|ccccccccc}
\hline

\multicolumn{1}{c|}{\multirow{3}{*}{Methods}} & \multicolumn{3}{c|}{\multirow{2}{*}{VeRi-776}} & \multicolumn{9}{c}{VehicleID} \\ \cline{5-13} 
 & \multicolumn{3}{c|}{} & \multicolumn{3}{c|}{Small} & \multicolumn{3}{c|}{Medium} & \multicolumn{3}{c}{Large} \\
 & mAP & R1 & R5 & R1 & R5 & \multicolumn{1}{c|}{mAP} & R1 & R5 & \multicolumn{1}{c|}{mAP} & R1 & R5 & mAP \\ \hline
PRN & 74.3 & 94.3 & 98.7 & 78.4 & 92.3 & \multicolumn{1}{c|}{-} & 75.0 & 88.3 & \multicolumn{1}{c|}{-} & 74.2 & 86.4 & - \\
SAN & 72.5 & 93.3 & 97.1 & 79.7 & 94.3 & \multicolumn{1}{c|}{-} & 78.4 & 91.3 & \multicolumn{1}{c|}{} & 75.6 & 88.3 &  \\
UMTS & 75.9 & 95.8 & - & 80.9 & - & \multicolumn{1}{c|}{87.0} & 78.8 & - & \multicolumn{1}{c|}{84.2} & 76.1 & - & 82.8 \\
PVEN & 79.5 & 95.6 & 98.4 & 84.7 & 97.0 & \multicolumn{1}{c|}{-} & 80.6 & 94.5 & \multicolumn{1}{c|}{-} & 77.8 & 92.0 & - \\
SAVER & 79.6 & 96.4 & 98.6 & 79.9 & 95.2 & \multicolumn{1}{c|}{-} & 77.6 & 91.1 & \multicolumn{1}{c|}{-} & 75.3 & 88.3 & - \\
CFVMNet & 77.1 & 95.3 & 98.4 & 81.4 & 94.1 & \multicolumn{1}{c|}{-} & 77.3 & 90.4 & \multicolumn{1}{c|}{-} & 74.7 & 88.7 & - \\
CAL & 74.3 & 95.4 & 97.9 & 82.5 & 94.7 & \multicolumn{1}{c|}{87.8} & 78.2 & 91.0 & \multicolumn{1}{c|}{83.8} & 75.1 & 88.5 & 80.9 \\ \hline
CLIP-ReID (CNN) & 80.3 & 96.8 & 98.4 & 85.2 & 97.1 & \multicolumn{1}{c|}{90.3} & 80.7 & 94.3 & \multicolumn{1}{c|}{86.5} & 78.7 & 92.3 & 84.6 \\
CLIP-ReID (ViT) & 83.3 & 97.4 & 98.6 & 85.3 & 97.6 & \multicolumn{1}{c|}{90.6} & 81.0 & 95.0 & \multicolumn{1}{c|}{86.9} & 78.1 & 92.7 & 84.4 \\ \hline
\end{tabular}
\caption{Comparisons with the state-of-the-art vehicle ReID methods on the VeRi-776 and VehicleID datasets.}
\label{tab:vehicleall}
\end{table*}

\end{document}